\documentclass{article}%[icg]{iosart2x}

\usepackage{subcaption}
\usepackage{booktabs,fontenc,graphicx,amsmath,url}

\newcommand{\reffigure}[1]{\figurename~\ref{#1}}
\newcommand{\refsection}[1]{Section~\ref{#1}}
\newcommand{\refsubsection}[1]{Subsection~\ref{#1}}
\newcommand{\reftable}[1]{Table~\ref{#1}}

\begin{document}
%\begin{frontmatter}

%\pretitle{}
\title{Deep Learning for General Game Playing with Ludii and Polygames}
%\runtitle{Deep Learning for General Game Playing with Ludii and Polygames}
%\title{Ludii: hundreds of games equipped with Zero Learning} % feel free to propose better :-)
%\runtitle{Ludii \& zero learning}
%\subtitle{}

% For one author:
%\author{\inits{N.}\fnms{Name1} {Surname1}\urle1]{first@somewhere.com}}
%\address{Department first, \orgname{University or Company name},
%Abbreviate US states, \cny{Country}\printead[presep={\\}]{e1}}

% Two or more authors:
\author{
{ {Dennis J. N. J.} {Soemers$*$}},
{{Vegard} {Mella$**$}},\\
{{Cameron} {Browne$*$}},
{{Olivier} {Teytaud$**$}}}

\maketitle

\begin{abstract}
Combinations of Monte-Carlo tree search and Deep Neural Networks, trained through self-play, have produced state-of-the-art results for automated game-playing in many board games. The training and search algorithms are not game-specific, but every individual game that these approaches are applied to still requires domain knowledge for the implementation of the game's rules, and constructing the neural network's architecture -- in particular the shapes of its input and output tensors. Ludii is a general game system that already contains over 500 different games, which can rapidly grow thanks to its powerful and user-friendly game description language. Polygames is a framework with training and search algorithms, which has already produced superhuman players for several board games. This paper describes the implementation of a bridge between Ludii and Polygames, which enables Polygames to train and evaluate models for games that are implemented and run through Ludii. We do not require any game-specific domain knowledge anymore, and instead leverage our domain knowledge of the Ludii system and its abstract state and move representations to write functions that can automatically determine the appropriate shapes for input and output tensors for any game implemented in Ludii. We describe experimental results for short training runs in a wide variety of different board games, and discuss several open problems and avenues for future research.\footnote{$*$Department of Data Science and Knowledge Engineering, Maastricht University, the Netherlands.$**$Facebook AI Research.}

\end{abstract}

%\begin{keyword}
%\kwd{General games}
%\kwd{Deep learning}
%\kwd{Ludii}
%\kwd{Polygames}
%\end{keyword}

%\end{frontmatter}

%%%%%%%%%%% The article body starts:

\section{Introduction} \label{Sec:Introduction}

Self-play training approaches such as those popularised by \textit{AlphaGo Zero} \cite{Silver2017AlphaGoZero} and \textit{AlphaZero} \cite{Silver2018AlphaZero}, based on combinations of Monte-Carlo tree search (MCTS) \cite{Kocsis2006BanditBased, Coulom2007, Browne2012} and Deep Learning \cite{LeCun2015}, have been demonstrated to be fairly generally applicable, and achieved state-of-the-art results in a variety of board games such as Go \cite{Silver2017AlphaGoZero}, Chess, Shogi \cite{Silver2018AlphaZero}, Hex, and Havannah \cite{Cazenave2020Polygames}. These approaches require relatively little domain knowledge, but still require some in the form of:
\begin{enumerate}
    \item A complete implementation of a forward model for the game, for the implementation of lookahead search as well as automated self-play to generate experience for training.
    \item Knowledge of which state features are required or useful to provide as inputs for a neural network.
    \item Knowledge of the action space, which is typically used to construct the policy head in such a way that every distinct possible action has a unique logit.
\end{enumerate}

The first requirement, for the implementation of a forward model, is partially addressed by research on using learned simulators for tree search as in \textit{MuZero} \cite{Schrittwieser2019MuZero}, but in practice a simulator is actually still required for the purpose of generating trajectories outside of the tree search. For the board games Go, Chess, and Shogi, MuZero still requires the input and output tensor shapes (for states and actions, respectively) to be manually designed per game. We remark that MuZero was also evaluated on 57 different Atari games in the Arcade Learning Environment (ALE) \cite{Bellemare2013ALE}, and it can use identical tensor shapes across all these Atari games because ALE uses the same observation and action spaces for all games in this framework.

The challenge posed by General Game Playing (GGP) \cite{Pitrat68GGP} is to build systems that can play a wide variety of games, which makes the three forms of required domain knowledge listed above difficult. A number of systems have been proposed that can interpret and run any arbitrary game as long as it has been described in their respective game description language, such as the original \textit{Game Description Language} (GDL) \cite{love08} from Stanford, \textit{Regular Boardgames} (RBG) \cite{kowalski19}, and \textit{Ludii} \cite{Piette2020Ludii}. 

In this paper, we describe how we combine the GGP system Ludii and the PyTorch-based \cite{Paszke2019PyTorch} state-of-the-art training algorithms in Polygames \cite{Cazenave2020Polygames}, with the goal of mitigating all three of the requirements for domain knowledge listed above. Section \ref{Sec:Background} provides some background information on these training techniques. Section \ref{Sec:DeepLearningGGP} describes existing work and limitations in applying these Deep Learning approaches to general games. Section \ref{ilp} presents the interface between Ludii and Polygames. Experiments and results are described in \refsection{Sec:Experiments}. We discuss some open problems in \refsection{Sec:OpenProblems}, and conclude the paper in \refsection{Sec:Conclusion}.

\section{Background} \label{Sec:Background}

\begin{figure}
    \centering
    \includegraphics[width=.95\textwidth]{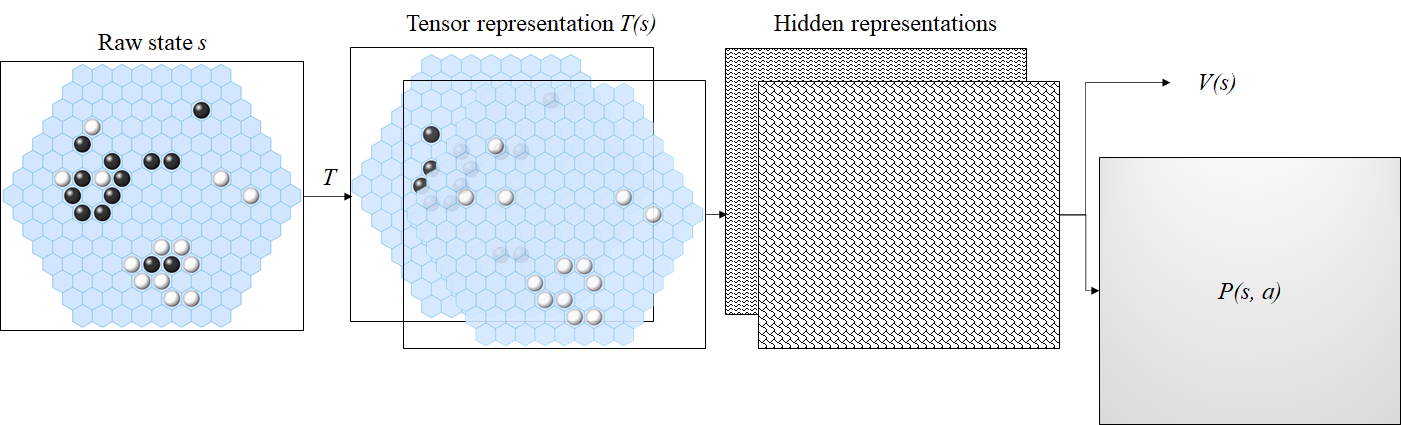}
    \caption{Basic architecture of DNNs for game playing. Raw game states $s$ are transformed into a tensor representation $T(s)$ of some fixed shape (often $3$-dimensional). The DNN learns to compute hidden representations of its inputs in hidden layers. Finally, it computes a scalar value estimate $V(s)$, and a discrete probability distribution $\mathbf{P}(s)$ with probabilities $P(s, a)$ for all actions $a$ in the complete action space.}
    \label{Fig:ExampleArchitecture}
\end{figure}

The basic premise behind AlphaZero and similar approaches in frameworks such as Polygames is that Deep Neural Networks (DNNs) take representations $T(s)$ of game states $s$ as input, and produce discrete probability distributions $\mathbf{P}(s)$ with probabilities $P(s, a)$ for all actions $a$ in states $s$, as well as value estimates $V(s)$, as outputs. This is depicted in \reffigure{Fig:ExampleArchitecture}. Both of these outputs are used to guide MCTS in different ways.

DNNs in general have a fixed architecture, requiring fixed and predetermined shapes for both the input and the output representations. The value output is always simply a scalar,\footnote{Assuming $2$-player zero-sum games; see \cite{Petosa2019MultiplayerAlphaZero} for relaxations of this assumption.} but determining the shapes of the input tensors $T(s)$ and policy outputs $\mathbf{P}(s)$ typically requires game-specific domain knowledge.

$T(s)$ is generally a $3$-dimensional tensor, where $2$ dimensions are spatial dimensions (corresponding to e.g. a $2$-dimensional playable area in a board game). The third dimension is formed by a stack of different channels which each have different semantics. For example, $T(s)$ in AlphaZero has a shape of $19$$\times$$19$$\times$$17$ for the game of Go played on a $19$$\times$$19$ board, with eight times two binary channels encoding the presence of the two players' pieces -- for a history of up to eight successive game states ending in $s$ -- and one final channel encoding the current player to move. The spatial structure of the first two dimensions is typically assumed to be meaningful, which is exploited by the inductive bias of Convolutional Neural Networks (CNNs) \cite{LeCun1989CNNs}.

For the policy head, it is customary for neural networks to first output real-valued \textit{logits} $L(s, a)$ for all possible actions $a$. These are subsequently converted into probabilities $P(s, a)$ using a softmax over all legal actions $a' \in \mathcal{A}(s)$ in $s$:
\begin{equation*}
    P(s, a) = \frac{\exp(L(s, a))}{\sum_{a' \in \mathcal{A}(s)} \exp(L(s, a'))}.
\end{equation*}
It is generally assumed that every distinct possible action $a$ that may be legal in any game state $s$ has a unique, matching logit $L(s, a)$. This means that domain knowledge of the game's action space is required to construct a DNN's architecture in such a way that distinct actions always have distinct logits. The logits are sometimes laid out in a structure of multiple $2$-dimensional planes, like the inputs, but typically preceded by fully connected (as opposed to convolutional) layers. This is equivalent to all the logits being laid out in a single, flat vector with no spatial structure.

\begin{figure}
\centering
\begin{subfigure}{.45\textwidth}
  \centering
  \includegraphics[width=\linewidth]{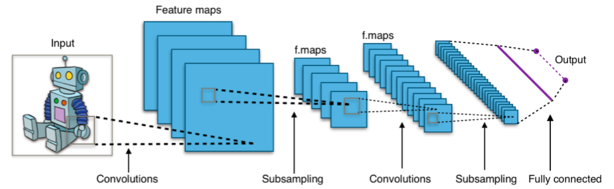}
  \caption{Standard convolutional net: convolutional layers first, with their inductive bias towards spatial invariance, followed by fully connected layers. (source: \url{https://en.wikipedia.org/wiki/Convolutional\_neural\_network\#/media/File:Typical\_cnn.png})}
  \label{Fig:StandardConvolutional}
\end{subfigure}\ \ \ 
\begin{subfigure}{.45\textwidth}
  \centering
  \includegraphics[width=\linewidth]{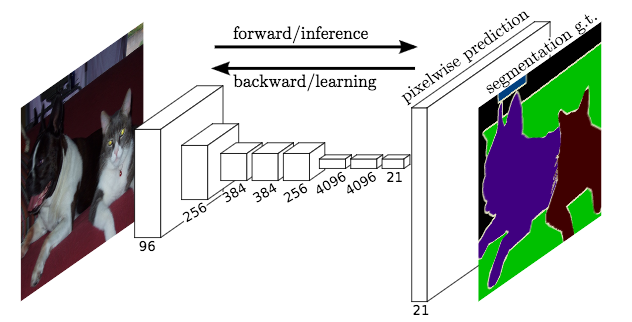}
  \caption{Fully convolutional net, e.g. for image segmentation: each output scalar (each logit in the case of games) is the output of the same net, applied on a moving window. No fully connected layers.}
  \label{Fig:FullyConvolutional}
\end{subfigure}\\
\begin{subfigure}{.45\linewidth}
  \centering
  \includegraphics[width=\linewidth]{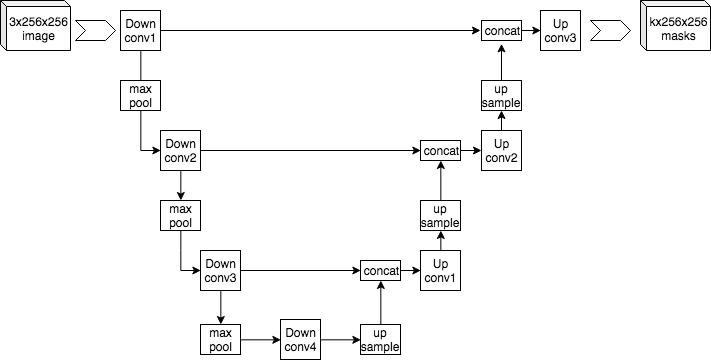}
  \caption{U-networks: this fully convolutional network also connects some layers to their symmetric counterpart supposed to work at a similar level of abstraction.}
  \label{Fig:UNet}
\end{subfigure}\ \ \ 
\begin{subfigure}{.45\textwidth}
  \centering
  \includegraphics[width=\linewidth]{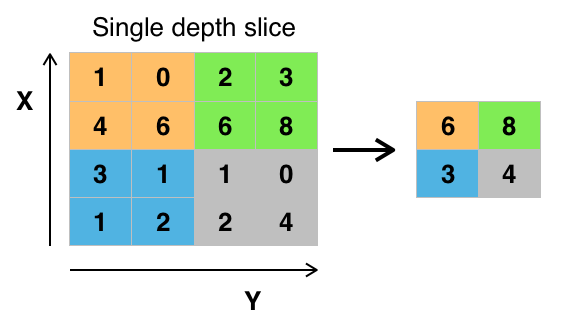}
  \caption{Max pooling: here we downsize from 4x4 to 2x2. In case of global pooling, the size of tensors after the global pooling layer is $1\times 1 \times \# channels$, independently of the input size.}
  \label{Fig:gp}
\end{subfigure}
\caption{Convolutional neural network (a). Fully convolutional counterpart (b, image from \cite{fc}; other images from Wikipedia), typically used in Image segmentation: image segmentation is related to policy heads in games in that the output has the same spatial coordinates at the input. U-networks (c): only convolutional layers, and skip connections symmetrically connecting layers. Global pooling (d): here we down-sample to a spatial size 1x1 in the value head: this is boardsize invariant. Global pooling can use channels for mean, standard deviation, max, etc: the number of channels is not necessarily preserved. (b+d) or (c+d) allow boardsize-invariant training \cite{Cazenave2020Polygames}.}
\label{Fig:Architectures}
\end{figure}

In addition to such typical architectures, Polygames \cite{Cazenave2020Polygames} includes various different structures, such as:
\begin{itemize}
    \item {\bf{Fully convolutional networks: }} As in many cases, actions are spatially distributed in a manner somehow close to the pieces, the output has spatial coordinates matching the spatial coordinates of the input. This can be exploited in fully convolutional networks \cite{fc}: the policy head has no fully connected layer, and directly maps inputs to outputs through convolutional blocks. This has the advantage of being boardsize invariant: we can train in size $13$$\times$$13$ and play in $19$$\times$$19$. Global pooling can be used to additionally make the value head size-invariant \cite{lin,greatfast}.
    \item {\bf{U-networks:}} It is usually considered that DNNs rephrase their data in an increasingly abstract manner, layer after layer. However, in fully convolutional networks, the output is dense; it has the same low-level nature as the input. The level of abstraction increases, and then decreases again. Then, one may consider that layers might benefit from a direct connection into a layer containing information at the same level of abstraction. This can be done by skip-connections, i.e. additional connections to layers symmetrically positioned in the network (\reffigure{Fig:UNet}): this is a U-network \cite{unets}.
\end{itemize}
Some of these different structures are depicted and explained in \reffigure{Fig:Architectures}.

\section{Deep Learning in General Game Playing} \label{Sec:DeepLearningGGP}

To some extent, all GGP systems mitigate the requirement for the implementation of complete forward models for every distinct game, in the sense that new games can be added and supported simply by defining them in a game description language. Ludii's game description language in particular has been designed in such a way that game descriptions for new games are fast and easy to write and understand \cite{Piette2020Ludii}, which has allowed for a significantly larger library of distinct games\footnote{Ludii has over 500 distinct built-in games at the time of this writing, with many of them having multiple variants for different board sizes, board shapes, variant rulesets, etc.} to be built up than would be feasible if they were all written in a programming language such as C++. Ludii's predecessor has also already demonstrated that the ``ludemic'' approach to game description languages used by Ludii facilitates procedural generation of complete games \cite{browne09}, which can be used to easily extend the set of compatible benchmark problems.

Similarly, we may argue that running games through a GGP system removes the requirements for \emph{game-specific knowledge} about how to shape state inputs and action outputs, but introduces requirements for similar \emph{knowledge about the GGP system}. Given any arbitrary game defined in a game description language of a GGP system, we require the ability to construct tensor representations of game states, and the ability to map from any index in a policy head to a matching action in any non-terminal game state.

GDL \cite{love08} is a low-level logic-based game description language, where games are described as logic programs consisting of many low-level propositions. Many GDL-based agents convert such a GDL description into a propositional network \cite{Schkufza2008Propositional,Cox2009Factoring,sironi17}, which can more efficiently process the games than Prolog-based reasoners or other similar techniques. Such propositional networks can be automatically constructed from GDL descriptions, and the structure of such a network remains constant across all game states of the same game. \cite{Goldwaser2020DeepRLGGP} therefore proposed using the internal state of a game's propositional network as the input state tensor for a deep neural network. A downside of this approach is that the state input tensor is a flat tensor, and there is no possibility to use inductive biases such as those of CNNs for inputs with spatial semantics. \textit{Galvanise Zero} \cite{Emslie2019GalvaniseZero} does exploit knowledge of spatial semantics through CNNs, but it only supports a limited selection of GDL-based games because it requires a handwritten Python function to create the mapping from game states to input tensors for every game that it supports. The action space can automatically be inferred from GDL descriptions, which means that these approaches require no extra domain knowledge with respect to the output policy heads.

In the game description language of Ludii \cite{Piette2020Ludii}, common high-level game concepts such as boards, piece types, etc. are all ``first-class citizen'' of the language, as opposed to GDL where every separate game description file encodes such concepts from scratch in low-level logic. Based on these concepts, Ludii also has an object-oriented game state representation that it uses internally, which remains consistent across all games. This enables us to write a single function that automatically constructs input tensors from Ludii's internal state representation, using our domain knowledge of Ludii as a whole instead of domain knowledge of every individual game. Unlike GDL, it is not straightforward (if at all possible) to infer the action space from game description files in Ludii. However, actions in Ludii do have an object-oriented structure, and at least an approximation of the action space can be constructed based on these properties -- again, based on domain knowledge of Ludii rather than any individual game. In many games, this is sufficient to distinguish most or all legal actions from each other.

\section{Interface Between Ludii and Polygames}\label{ilp}
Based on the insights described above, we developed an interface between the Ludii general game system, and the Polygames framework with state-of-the-art AI training code. In Polygames, different games are normally implemented from scratch in C++. The basic idea of this interface is that there is a single ``Ludii game'' in Polygames, with C++ code that interacts with Ludii's Java-based API through Java Native Interface. Polygames command-line arguments can be used to load different games and variants from Ludii into this wrapper. This section provides details on how Ludii automatically constructs tensor representations of its state and action spaces, based on its own internal representations, for any arbitrary game implemented in Ludii.\footnote{The source code for building tensors from Ludii's internal state and action representations is available from \url{https://github.com/Ludeme/LudiiAI}. All the source code of Polygames is available from \url{https://github.com/facebookincubator/Polygames}.} 

\subsection{Constructing the Spatial Dimensions} \label{Subsec:LudiiSpatialDimensions}

CNNs normally operate on grid structures of ``pixels'', such that every position can be indexed by a row and column, and every position has a square of up to eight neighbour positions around it. This structure resembles the game boards of games such as Chess, Shogi, and Go most closely. Some other boards, such as the tilings of hexagonal cells used in games like Hex and Havannah, can also be ``packed'' into such a grid. This approach is used for the game-specific C++ implementations of those games in Polygames. However, Ludii supports games with arbitrary graphs as boards, and hence requires a generic solution that can map positions from graphs with any arbitrary connectivity structure into a grid structure that CNNs can work with. 

For every game in Ludii, there is at least one (and possibly more than one) \textit{container}, which specifies a playable ``area'' with positions that may contain pieces, have corresponding clickable elements in Ludii's GUI, etc. \cite{Piette2020Ludii,Piette2020LudiiGameLogicGuide}. The first container typically corresponds to the board that a game is played on, and is often the largest. Any other containers represent auxiliary areas, such as players' hands to hold captured pieces in Shogi. Even games that are not generally thought of as being played on a board are still modelled in this way in Ludii. For instance, Rock-Paper-Scissors is modelled as a board with two (initially empty) cells, and two hands for the two players, each containing rock, paper, and scissors ``pieces'' which players can drag onto their designated cells on the board to make their move.

\begin{figure}[t]
\centering
\begin{subfigure}{.5\textwidth}
  \centering
  \includegraphics[width=.63\linewidth]{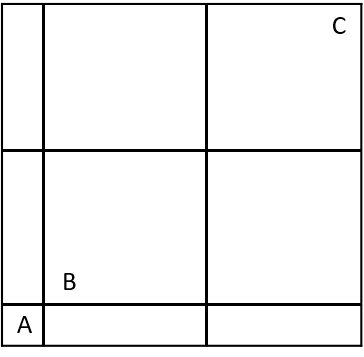}
  \caption{A $3$$\times$$3$ grid is computed based on the distinct $x$-\\ and $y$-coordinates of three sites A, B, and C.}
  \label{Fig:LudiiGrid}
\end{subfigure}%
\begin{subfigure}{.5\textwidth}
  \centering
  \includegraphics[width=.63\linewidth]{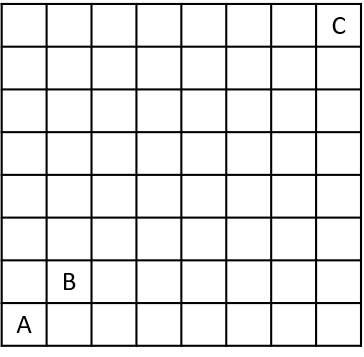}
  \caption{An $8$$\times$$8$ grid is overlaid on the $[0, 1]^2$ space containing three sites A, B, and C.}
  \label{Fig:FullGrid}
\end{subfigure}
\caption{Two different approaches for computing a grid based on a playable space defined by three sites A, B, and C, each with distinct $x$- and $y$-coordinates. The approach we use is depicted in (a). This approach results in smaller, more dense tensors, but information of the relative distances between all sites is not necessarily preserved. The alternative approach, depicted in (b), preserves more of this information, but can result in large and sparse tensors.}
\label{Fig:Grids}
\end{figure}

Every site in any such container in Ludii has $x$ and $y$ coordinates in $[0, 1]$, which are used by Ludii for purposes such as drawing game states in the GUI for human players. We construct a grid structure simply by sorting all the distinct $x$- and $y$-coordinates across all sites in the board in increasing order, and assigning distinct columns and rows, respectively, to distinct $x$- and $y$-coordinates. Coordinates that are within a tolerance value of $10^{-5}$ are treated as equal, to avoid generating excessively large and sparse tensors due to small differences resulting from floating-point arithmetic. Note that this approach is not equivalent to directly overlaying a sufficiently fine-grained grid over the $[0, 1]^2$ space, because we only add rows and columns that each contain at least one site. This is depicted in \reffigure{Fig:Grids}. Our approach may lose some information concerning the relative distances between sites, but because these $x$- and $y$-coordinates are only used for the graphical user interface -- not for game logic -- we expect the smaller and less sparse grids to be preferable due to improved computational efficiency. Note that the vast majority of games in Ludii use boards defined by regular or semiregular tilings, and for these the two approaches will have similar results.

\begin{figure}
\centering
  \includegraphics[width=.95\textwidth]{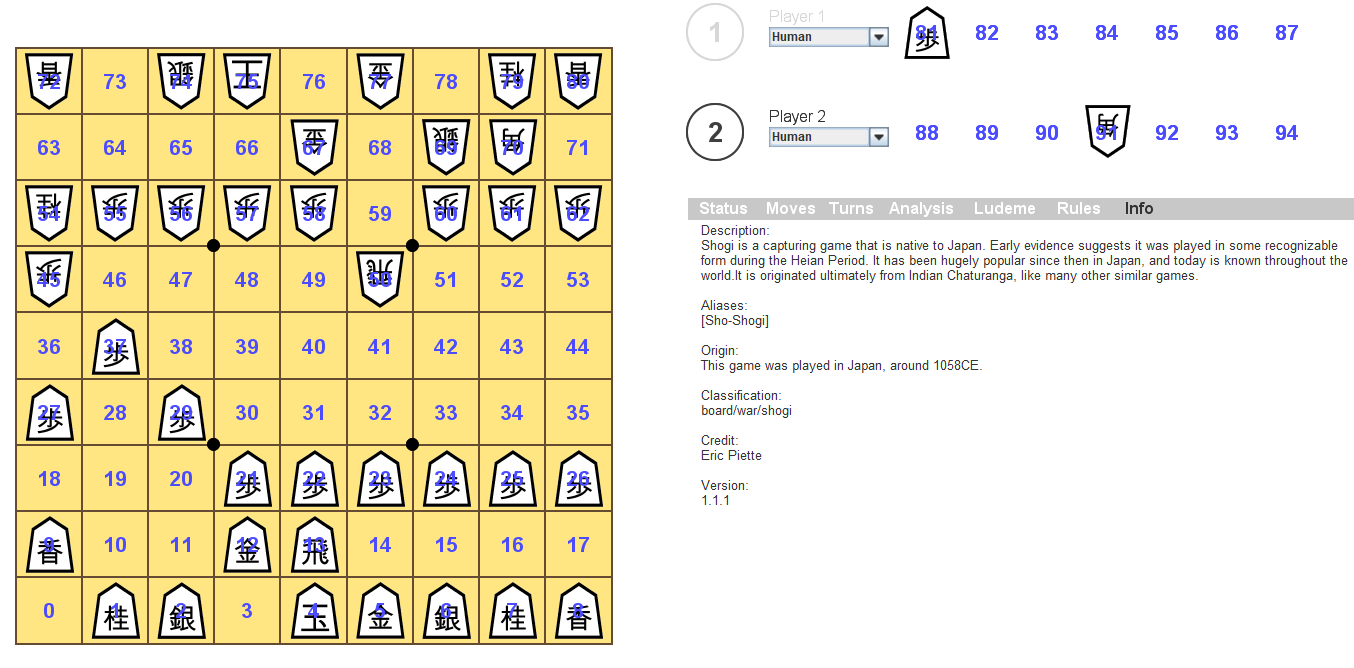}
  \caption{Shogi being played in Ludii's user interface. The game board is on the left-hand side, and each player has a ``hand'' with seven slots to hold captured pieces on the right-hand side. \reffigure{Fig:ShogiLudiiTensorShape} shows how the numbered positions get mapped to positions in a tensor.}
  \label{Fig:ShogiLudii}
\end{figure}

\begin{figure}
  \centering
  \includegraphics[width=.6\textwidth]{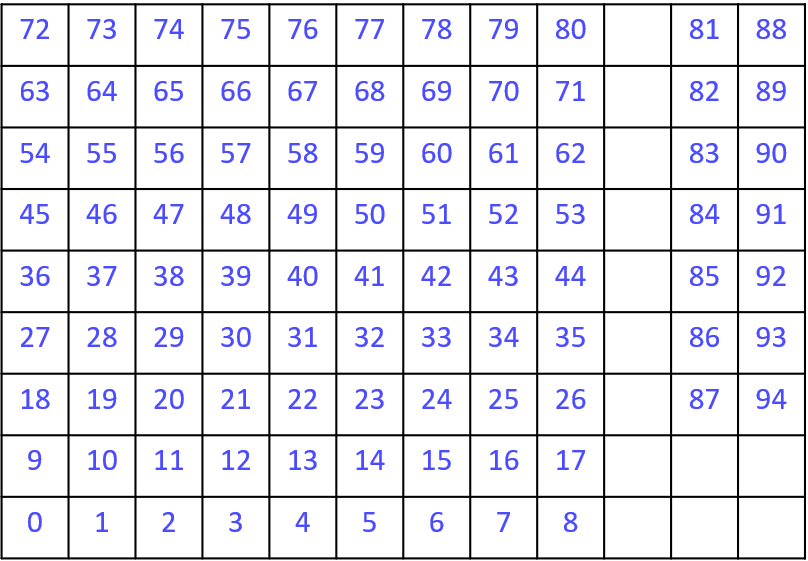}
  \caption{Mapping from positions in Shogi's three containers to positions in a single tensor. Numbers $0$ through $80$ correspond to positions on the board, $81$ through $87$ are positions in the hand of Player 1, and $88$ through $94$ are positions in the hand of Player 2 (see \reffigure{Fig:ShogiLudii}).}
  \label{Fig:ShogiLudiiTensorShape}
\end{figure}

In the current version of Ludii, containers other than the first one (corresponding to the ``main'' board) never have more than one meaningful dimension; they are always a single, contiguous sequence of cells. Each of those containers is concatenated to the grid constructed for the first container, either using one extra column or one extra row per extra container (whichever results in the lowest increase in total size of the tensor). Additionally, one extra dummy row or column is inserted to create a more explicit separation between the main board (for which we expect there to be meaningful spatial semantics) and the other containers (for which there is no expectation that any meaningful spatial semantics exist). For example, Shogi is played on a $9$$\times$$9$ board, but each of the two players also has a ``hand'' of $7$ cells as extra containers to potentially hold captured pieces. This results in a $12$$\times$$9$ grid for Shogi. A screenshot of Shogi being played in Ludii's user interface is depicted in \reffigure{Fig:ShogiLudii}, with cells of the different containers labelled by numbers. The mapping from these positions to positions in the tensor representation is depicted in \reffigure{Fig:ShogiLudiiTensorShape}.

\subsection{Representing Ludii Game States as Tensors} \label{Subsec:LudiiStateTensors}

Let $s$ denote a raw game state in Ludii's object-oriented state representation \cite{Piette2020LudiiGameLogicGuide}, for a game $\mathcal{G}$. Based on the properties of $s$, we construct a tensor representation $T(s)$ -- which can be used as input for a DNN -- of shape $(C, W, H)$, where $C$ denotes the number of channels (variable, depends on $\mathcal{G}$), $W$ denotes the width (i.e., number of columns), and $H$ denotes the height (i.e., number of rows). The channels are constructed as follows:
\begin{itemize}
    \item Binary channels indicating the presence (or absence) of every piece type defined in $\mathcal{G}$. Most games have one channel per piece type, where values of $1$ indicate the presence of a piece of that type in a position. If $\mathcal{G}$ is a ``stacking'' game, meaning that it allows for multiple pieces to form a stack on a single position, we use $M + N$ binary channels per piece type, instead of just one. $M$ channels are used to indicate presence of a piece type in the bottom $M$ layers of every stack on every position, and $N$ channels indicate the same for the top $N$ layers. In our implementation, we use $M = N = 5$. If a single stack contains more than $M + N$ pieces, this representation is not sufficient to provide information about some of the middle layers to the DNN, but this is rare in practice.
    \item If $\mathcal{G}$ is a ``stacking'' game, we include an additional non-binary channel containing the height of every stack in every position.
    \item If $\mathcal{G}$ is a game where positions can contain a ``count'' of more than one piece, we include a non-binary channel denoting the count of pieces on that position. This channel is semantically similar to the one described above for stack heights. In Ludii, positions in these games are still restricted to containing only a single piece \textit{type} at a time. This is most notably used for \textit{mancala} games. Games where pieces of different types can share a single position are modelled as stacking games instead.
    \item Ludii's state representation can include an ``amount'' value per player, primarily intended to represent money for games that involve betting or other similar mechanisms. If $\mathcal{G}$ uses this, we add one non-binary channel per player, such that every position in the channel for player $p$ contains the amount value of $p$ in $s$.
    \item If $\mathcal{G}$ is played by $n > 1$ players, we include $n$ binary channels, such that the $n^{th}$ channel is filled with values of $1$ if and only if $n$ is the current player to move in state $s$. This also accounts for swap rules. For example, the first player normally plays red, and the second blue, in \textit{Hex}. If $s$ is a state where the red player is the next to make a move, and a swap has occurred, the second of these channels will be filled with $1$ entries instead of the first.
    \item In some games, every position has a ``local state'' variable, which is an integer value. Different games can use this in different ways to store (temporary) auxiliary information about positions. For instance, local state values of $1$ are used for positions that contain pieces that are still in their initial position, and values of $0$ otherwise (this is used for castling). Most games only use low local state values, if any at all. Hence, we use separate binary channels to indicate local state values of $0$, $1$, $2$, $3$, $4$, and $\geq 5$.
    \item If the game uses a swap rule (or ``pie rule''), such as Hex, we include a binary channel that is filled with values of $1$ if and only if a swap has occurred in $s$.
    \item For every distinct container in $\mathcal{G}$, we include one binary channel that has values of $1$ for entries that correspond to a position in that container, and values of $0$ everywhere else.
    \item For each of the last two moves $m$ played prior to reaching $s$, we add one binary channel with only a single value of $1$ in the entry corresponding to the ``from'' position of $m$ (typically the location that a piece moves away from), and a similar channel for the ``to'' position of $m$ (typically the location that a piece is placed in).
\end{itemize}

With these channels, we did not yet exhaustively cover all the state variables in Ludii's game state representation \cite{Piette2020LudiiGameLogicGuide}, but we covered the most commonly-used ones. Whenever new variables are added to Ludii's game state representation, engineering effort for including these in the tensor representations is only required once for Ludii as a whole -- not once per game added to Ludii.

\subsection{Representing Ludii Actions as Tensors} \label{Subsec:LudiiActionTensors}

In contrast to GDL \cite{love08,Emslie2019GalvaniseZero,Goldwaser2020DeepRLGGP}, it is not straightforward -- if at all possible -- to automatically infer the complete action space for any arbitrary game described in Ludii's game description language. This is because in Ludii's game description language, the function that generates lists of legal moves is defined as a composite of many simple functions (\textit{ludemes}), which may be arranged in any arbitrary tree structure. While each of these ludemes in principle has some domain for its possible inputs, and range for its possible outputs, these are not strictly defined in logic-based or other formats that permit automated inference.

Similar to its state representation, Ludii has an object-oriented move\footnote{In this document we use the terms ``move'' and ``action'' interchangeably, to refer to complete decisions that players make. Within Ludii, these are referred to only as moves, and actions are smaller parts of moves.} representation \cite{Piette2020LudiiGameLogicGuide}. However, in contrast to the state representation, the most important variables of the move representation are arbitrarily-sized lists (of primitive modifications to be applied to a game state) and arbitrarily-sized trees (of ludemes to be evaluated after applying the initial primitive modifications). The arbitrary sizes of these variables make them difficult to encode in a fixed-size tensor representation. Hence, we ignore these properties, and only distinguish moves based on some simple properties that can easily be used for this purpose. We construct the space of output tensors to map moves to for a game $\mathcal{G}$ as follows:
\begin{itemize}
    \item The action space is organised as a stack of $2$-dimensional planes, with the spatial dimensions being identical to those of the state tensors (see \refsubsection{Subsec:LudiiStateTensors}). Every action will map to exactly one position in this space -- i.e., one location in the $2$-dimensional area, and one channel.
    \item Pass and swap moves have been identified as special cases that are sufficiently common, important, and semantically different from any other kind of move that they warrant the inclusion of their own dedicated channels. 
    \item Many games only involve moves that can be identified by just a single position in the spatial dimensions; these are generally games where players place stones (Go, Hex, Havannah, etc.), but may in theory also be games like Chess if they have been defined in a way such that movements are split up into two separate decisions (picking a source and picking a destination). These games can be automatically discovered in Ludii. For these games, we only add one more channel in addition to the pass and swap move channels, to encode all other moves based on their positions in the spatial dimensions. In Ludii, this position is referred to as the ``to'' position.
    \item In all other games, moves may have distinct ``from'' and ``to'' positions; typical examples are standard implementations of Chess, Amazons, Shogi, etc. For moves that have an invalid ``from'' position, we assume that it is equal to the ``to'' position. For games that involve stacking, moves may additionally have $l_{min}$ and $l_{max}$ properties which refer to the levels within a stack at which a move operates; both are assumed to equal $0$ if the game does not allow stacking. The ``to'' position of a move is used to map the move to a location in the spatial dimensions, and the remaining properties are used to index into one of multiple channels based on the relative ``distance covered'' by the move. More specifically, we create $(2M + 1) \times (2M + 1) \times (N + 1) \times (N + 1)$ channels, where we use $M = 3$, and $N = 2$ if $\mathcal{G}$ involves stacking, or $N = 0$ otherwise. Let $dx$ and $dy$ denote the differences in rows and columns, respectively, between the ``to'' and ``from'' positions of a move. Let $[a]_b^c$ denote a value of $a$ clipped to lie in the interval $[b, c]$. Then, this move gets mapped to the channel given by the $0$-based index $\left(\left([dx]_{-M}^M \times (2M + 1) + [dy]_{-M}^M\right) \times (N + 1) + [l_{min}]_0^N\right) \times (N + 1) + [l_{max} - l_{min}]_0^N$.
\end{itemize}
Note that this is simply one approach to constructing tensor representations of moves that we implemented, but we may envision other approaches as well. For instance, in a game like Chess, it may be more important to encode the type of the piece that makes a move, rather than encoding the distance and direction covered by a move. This could be accomplished by creating channels that are indexed based on the type of piece in the ``from'' location of a move, instead of the distance between ``from'' and ``to'' positions.

While we find this approach to be sufficient to distinguish moves from each other in many cases, there are cases where multiple distinct moves that are legal in a single game state will end up being represented by exactly the same logit. When multiple distinct moves are represented by the same logit in a DNN's output, we say that they are \textit{aliased}. DNNs cannot distinguish between aliased moves, and hence always provide the same advice (in the form of the prior probabilities $P(s, a)$) to MCTS for these different moves. However, in Polygames \cite{Cazenave2020Polygames}, the MCTS itself \emph{can} still distinguish between the different moves by different representing them as distinct branches in the search tree, and backing up (potentially) different values throughout the tree search. This is an important difference with other frameworks, such as OpenSpiel \cite{LanctotEtAl2019OpenSpiel}, where the MCTS itself requires every possible distinct action that may ever be legal in a game to be assigned a unique integer upfront. When subsequently using the visit counts to compute the standard cross-entropy loss as proposed by \cite{Silver2017AlphaGoZero}, the visit counts for all moves that share a single logit are summed up. The softmax over the logits only counts every distinct logit once.

\section{Experiments} \label{Sec:Experiments}

\begin{figure}
\centering
\begin{subfigure}{.19\textwidth}
  \centering
  \includegraphics[width=\linewidth]{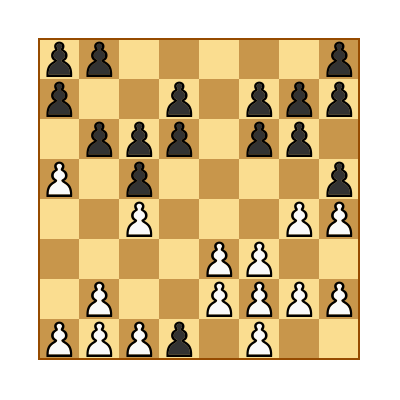}
\end{subfigure}
\begin{subfigure}{.19\textwidth}
  \centering
  \includegraphics[width=\linewidth]{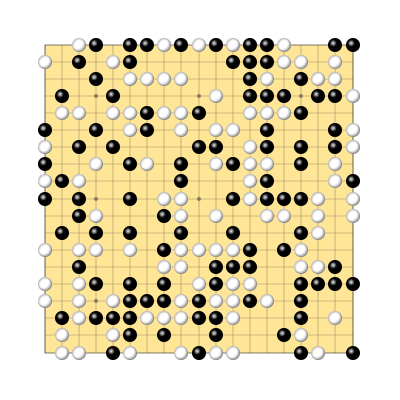}
\end{subfigure}
\begin{subfigure}{.19\linewidth}
  \centering
  \includegraphics[width=\linewidth]{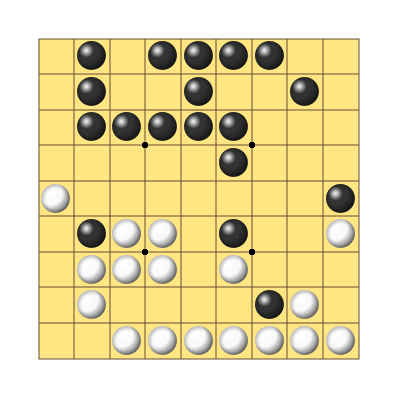}
\end{subfigure}
\begin{subfigure}{.19\textwidth}
  \centering
  \includegraphics[width=\linewidth]{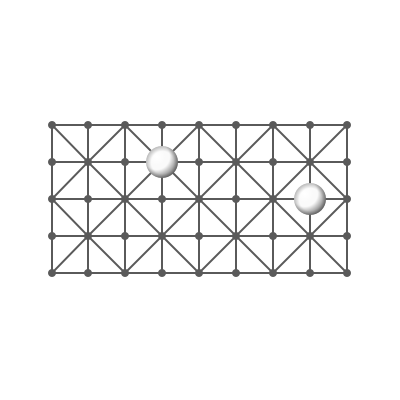}
\end{subfigure}
\begin{subfigure}{.19\textwidth}
  \centering
  \includegraphics[width=\linewidth]{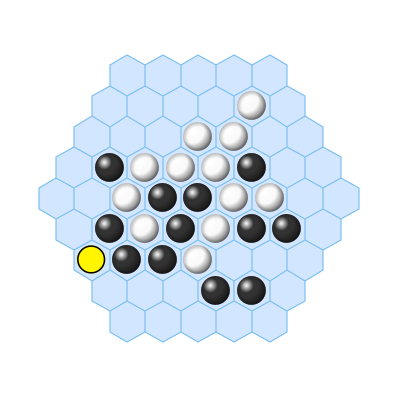}
\end{subfigure}
\begin{subfigure}{.19\textwidth}
  \centering
  \includegraphics[width=\linewidth]{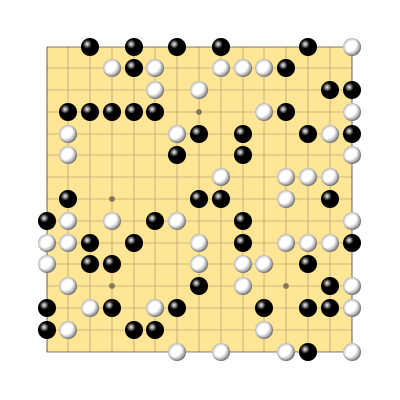}
\end{subfigure}
\begin{subfigure}{.19\textwidth}
  \centering
  \includegraphics[width=\linewidth]{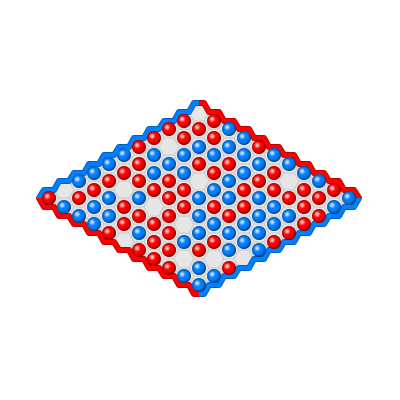}
\end{subfigure}
\begin{subfigure}{.19\textwidth}
  \centering
  \includegraphics[width=\linewidth]{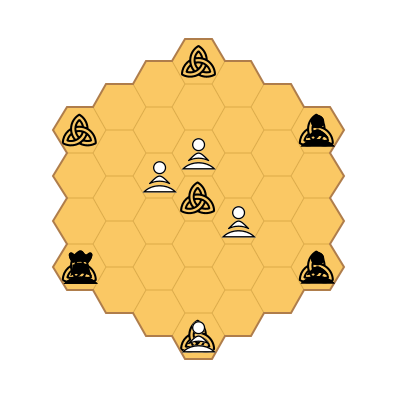}
\end{subfigure}
\begin{subfigure}{.19\textwidth}
  \centering
  \includegraphics[width=\linewidth]{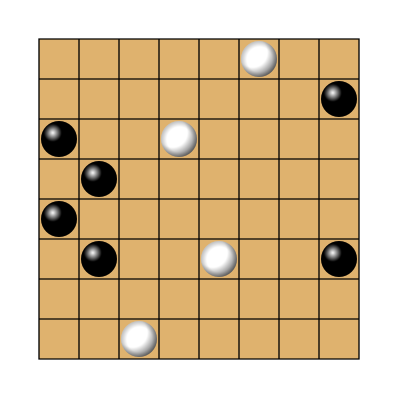}
\end{subfigure}
\begin{subfigure}{.19\textwidth}
  \centering
  \includegraphics[width=\linewidth]{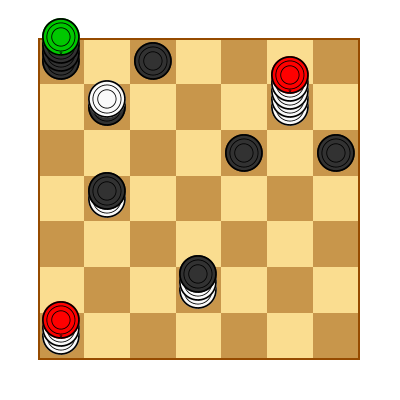}
\end{subfigure}
\begin{subfigure}{.19\textwidth}
  \centering
  \includegraphics[width=\linewidth]{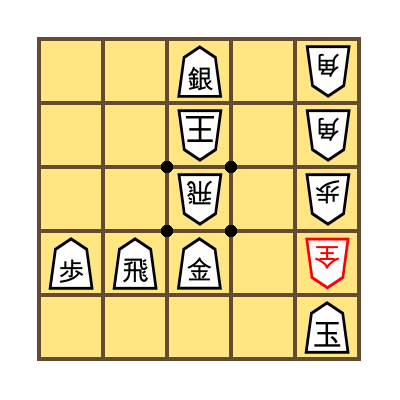}
\end{subfigure}
\begin{subfigure}{.19\textwidth}
  \centering
  \includegraphics[width=\linewidth]{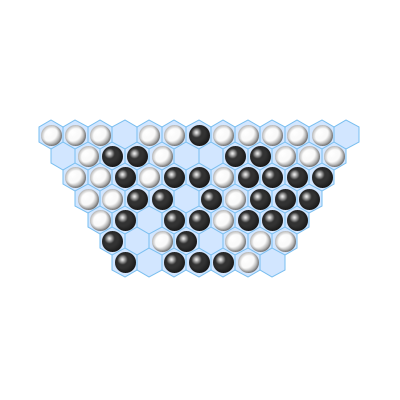}
\end{subfigure}
\begin{subfigure}{.19\textwidth}
  \centering
  \includegraphics[width=\linewidth]{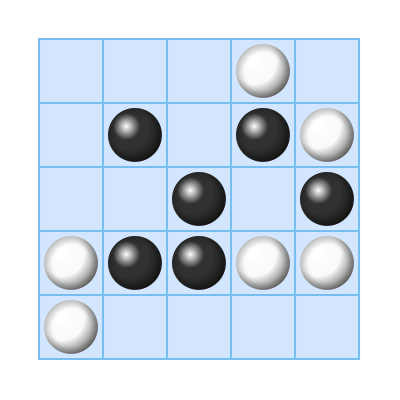}
\end{subfigure}
\begin{subfigure}{.19\textwidth}
  \centering
  \includegraphics[width=\linewidth]{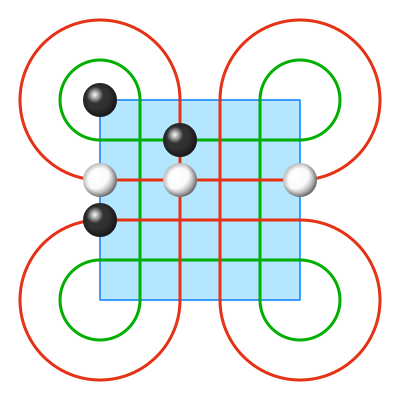}
\end{subfigure}
\begin{subfigure}{.19\textwidth}
  \centering
  \includegraphics[width=\linewidth]{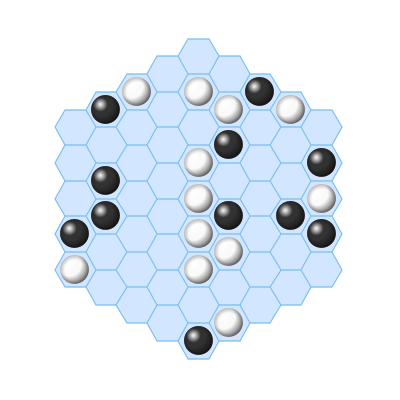}
\end{subfigure}
\caption{Screenshots of all the Ludii-based games included in our experiments. \textbf{First row}: Breakthrough, Connect6, Dai Hasami Shogi, Fanorona, Feed the Ducks. \textbf{Second row}: Gomoku, Hex, HeXentafl, Konane, Lasca. \textbf{Third row}: Minishogi, Pentalath, Squava, Surakarta, Yavalath.}
\label{Fig:GameThumbnails}
\end{figure}

In this section we describe experiments\footnote{The code used by Ludii to construct state and move tensors for any game is available from \url{https://github.com/Ludeme/LudiiAI}. All the training and evaluation code of Polygames is available from \url{https://github.com/facebookincubator/Polygames}. Checkpoints of models used in these experiments are available from \url{http://dl.fbaipublicfiles.com/polygames/ludii_checkpoints/list.txt}.} intended to demonstrate the potential for the approach described in the previous section to facilitate training and research in general games. We picked fifteen different games, all as implemented with their default options in Ludii \cite{Piette2020Ludii} v1.1.6, and trained a model of the \texttt{ResConvConvLogitPoolModelV2} type from Polygames \cite{Cazenave2020Polygames} in each of these games. The selected games are depicted in \reffigure{Fig:GameThumbnails}. 

We used the same training hyperparameters across all games. Every training run used 20 hours of wall time, with 8 GPUs, 80 CPU cores, and 475GB memory allocated per training job. Every training job used 1 server for model training, and 7 clients for the generation of self-play games. The MCTS agents used 400 MCTS iterations per move in self-play. 

The final model checkpoint of every training run is evaluated in a set of 300 evaluation games played against a pure MCTS -- a standard UCT agent \cite{Kocsis2006BanditBased,Browne2012} without any DNNs. In evaluation games, the MCTS with a trained model used 40 iterations per move, whereas the pure MCTS used 800 iterations per move -- where at the end of every iteration, the average outcome of 10 random rollouts is backed up. The final column of \reftable{Table:Results} reports the win percentages of the trained MCTS against the untrained MCTS. The table also provides further details on the number of trainable parameters in each of the DNNs, and for some games summarises unusual properties that these games have which we did not yet observe in much of the existing literature on learning through self-play in games.

\begin{table}
\caption{Data for a variety of different games, all implemented in Ludii v1.1.6, for which we trained models in Polygames over a duration of 20 hours using 8 GPUs and 80 CPU cores per model. The second column lists some interesting properties for games that we have not yet often seen (if at all) in existing literature using AlphaZero-like training approaches. The third column lists the number of trainable parameters in the model (we used identical Polygames hyperparameters for the DNN architecture across all games, but in Polygames by default the number of channels in hidden convolutional layers scales with the number of input channels). The last column lists the win percentages of MCTS with the trained model using 40 iterations per move, against MCTS without any trained model using 800 iterations per move -- where every iteration backs up the average outcome of 10 random rollouts. }
\label{Table:Results}
\vspace{6pt}
\centering
\scriptsize
%\begin{tabular}{@{}l|lrr@{}}
\begin{tabular}{|p{2.2cm}|p{3.8cm}|p{1.3cm}|p{1.3cm}|}
\toprule
\textbf{Game} & \textbf{Unusual Properties} & \textbf{Trainable Parameters} & \textbf{Win Percentage} \\
\midrule
Breakthrough & - & 188,296 & 100.00\% \\
Connect6 & - & 180,472 & 75.67\% \\
Dai Hasami Shogi & - & 188,296 & 99.33\% \\
Fanorona & Move aliasing due to choice of capture direction. & 188,296 & 50.00\% \\
Feed the Ducks & Moves have global effects across entire board. & 231,152 & 83.00\% \\
Gomoku & - & 180,472 & 91.00\% \\
Hex & - & 222,464 & 100.00\% \\
HeXentafl & Asymmetry in piece types, initial setup, and goals. & 231,152 & 98.67\% \\
Konane & - & 188,296 & 98.00\% \\
Lasca & Pieces (of multiple different types) can stack. & 5,450,268 & 3.50\% \\
Minishogi & - & 2,009,752 & 97.00\% \\
Pentalath & - & 180,472 & 95.33\% \\
Squava & Lines of $4$ win, but lines of $3$ lose. & 222,464 & 96.67\% \\
Surakarta & Loops around board allow for unique move patterns. & 188,948 & 100.00\% \\
Yavalath & Lines of $4$ win, but lines of $3$ lose. & 222,464 & 97.33\% \\
\bottomrule
\end{tabular}
\end{table}

In the majority of the evaluated games, the trained MCTS easily outperforms the untrained one, even using 20 times fewer MCTS iterations (or 200 times fewer if the number of random rollouts performed by the untrained MCTS is counted). Note that, in comparison to work that focuses on achieving superhuman playing strength \cite{Silver2018AlphaZero, Cazenave2020Polygames}, we focused on short training runs using fewer resources and smaller networks. Our primary aim is to demonstrate the possibility of training effectively using a single implementation without game-specific domain knowledge. 

The two results that stand out most are for \textit{Lasca} and \textit{Fanorona}. The win percentage of $3.50\%$ for Lasca indicates that this model is not trained nearly as well as the others. Lasca is the only game among those tested that involves stacking of multiple pieces on a single site. Our procedures for the construction of input and output tensors lead to a significantly larger numbers of channels in this game compared to the other games, which is also reflected in the large number of trainable parameters that this model has. Further research is required to establish whether it would be sufficient to reduce the size of the model, or whether entirely different approaches for constructing the tensors would be more appropriate. In Fanorona, the win percentage of $50\%$ for the trained model is not necessarily a poor level of performance (considering the large difference in number of MCTS iterations), but it appears to be noticeably worse than in the other games. One possible explanation for this may be that Fanorona has a more severe degree of move aliasing, because there are situations where there are multiple different legal moves with identical ``to'' and ``from'' positions, but different effects in that a player can choose in which direction they wish to capture opposing pieces. Such moves are all represented by a single, shared logit in our output tensors -- which means that only the MCTS can distinguish between them, but the trained policy head cannot.

\section{Open Problems} \label{Sec:OpenProblems}

Thanks to the large library of games available in Ludii \cite{Piette2020Ludii}, we can get a clear picture of categories of games that are open problems to various extents; some that are simply not supported yet by Polygames \cite{Cazenave2020Polygames} and require more engineering effort, and some that appear to have been neglected across the majority of recent game AI literature. All of these types of games are supported by Ludii:
\begin{itemize}
    \item \textit{Stochastic games}: these were not included in this paper because they are temporarily unsupported by the MCTS implementation of Polygames, but were supported in earlier versions of Polygames and will be again in future versions.
    \item \textit{Games with more than $2$ players}: support for these can be added relatively easily \cite{Petosa2019MultiplayerAlphaZero}, but is not yet available in Polygames.
    \item \textit{Imperfect-information games}: there has been some recent work towards AlphaZero-like training approaches that support imperfect-information games \cite{Brown2020Combining}, but tractability is still a concern for games with little common knowledge.
    \item \textit{Simultaneous-move games}: simultaneous-move games will at least require significant changes in the MCTS component \cite{Browne2012} as it is typically used in AlphaZero-like training setups.
    \item \textit{Games with excessively large state or move tensors}: games such as \textit{Taikyoku Shogi}, with a $36$$\times$$36$ board and $402$ pieces per player of $209$ different types, can be modelled and run in Ludii, but produce excessively large tensors which quickly lead to memory issues when training with standard hyperparameter values that work well for ``normal'' games. These issues do not appear straightforward to resolve with current hardware and large DNNs.
    \item \textit{Games played on a mix of cells, edges and/or vertices of graphs}: while games like \textit{Chess} are only played on cells, and games like \textit{Go} only on vertices, there are also games such as \textit{Contagion} that are played on a mix of multiple different parts of a graph. It is not clear how to directly support these with the standard CNNs.
    \item \textit{Games without an explicitly defined board}: games such as \textit{Andantino} or \textit{Chex} are not played in a limited area that is defined upfront, but in a playable area that grows dynamically as play progresses. The standard DNN architectures require these spatial dimensions to be predefined and fixed.
    \item \textit{Games with more than $2$ spatial dimensions}: games such as \textit{Spline} have a third spatial dimension, which cannot be handled by the standard $2$D convolutional layers. While a straightforward extension to $3$D convolutional layers may be sufficient, we are not aware of any existing research towards this for games, and also imagine that a third spatial dimension can rapidly lead to tensors becoming excessively large again for many non-trivial games.
\end{itemize}

\section{Conclusions} \label{Sec:Conclusion}

We have described our approach for constructing tensor representations of states and moves for any game implemented in the Ludii general game system, and used this to implement a bridge between Ludii and the Polygames framework. This allows for the state-of-the-art tree search and self-play training techniques implemented in Polygames to be used for training game-playing models in any game described in Ludii's general game description language. Whereas AlphaZero-like approaches typically require game-specific domain knowledge to define a Deep Neural Network's architecture and its input and output tensors, we only require such domain knowledge at the level of the general game system as a whole, and can now leverage Ludii's wide library of games -- which can quickly grow thanks to its user-friendly game description language -- to facilitate more general game AI research with minimal requirements for game-specific engineering efforts. We have identified a series of ``open problems'' in the form of classes of games that are already supported by Ludii, but not yet by Polygames. For some of these is a clear path that merely requires additional engineering effort, but others are likely to require a more significant amount of extra research.

%\begin{figure}[t]
%\caption{Figure caption.}\label{f1}
%\end{figure}

%\begin{table*}
%\caption{} \label{t1}
%\begin{tabular}{lll}
%\hline
%&&\\
%&&\\
%\hline
%\end{tabular}
%\end{table*}

%%%%%%%%%%% The bibliography starts:

%%  ios2_nameyear.bst will be used to                      %%

\section*{Acknowledgments}

This work was partially supported by the European Research Council as part of the Digital Ludeme Project (ERC Consolidator Grant \#771292), led by Cameron Browne at Maastricht University’s Department of Data Science and Knowledge Engineering. We thank {\'E}ric Piette for his image editing skills, and Matthew Stephenson for his mastery of the English language.

% if your bibliography is in bibtex format, use those commands:
\bibliographystyle{abbrv}  % Style BST file.
\bibliography{References,z,rl}        % Bibliography file (usually '*.bib')

\begin{thebibliography}{10}

\bibitem{Bellemare2013ALE}
M.~G. Bellemare, Y.~Naddaf, J.~Veness, and M.~Bowling.
\newblock {T}he {A}rcade {L}earning {E}nvironment: {A}n {E}valuation {P}latform
  for {G}eneral {A}gents.
\newblock {\em Journal of Artificial Intelligence Research}, 47:253--279, 2013.

\bibitem{Brown2020Combining}
N.~Brown, A.~Bakhtin, A.~Lerer, and Q.~Gong.
\newblock Combining deep reinforcement learning and search for
  imperfect-information games.
\newblock In H.~Larochelle, M.~Ranzato, R.~Hadsell, M.~Balcan, and H.~Lin,
  editors, {\em Advances in Neural Information Processing Systems 33 (NeurIPS
  2020)}, 2020.

\bibitem{Browne2012}
C.~Browne, E.~Powley, D.~Whitehouse, S.~Lucas, P.~I. Cowling, P.~Rohlfshagen,
  S.~Tavener, D.~Perez, S.~Samothrakis, and S.~Colton.
\newblock A survey of {M}onte {C}arlo tree search methods.
\newblock {\em IEEE Transactions on Computational Intelligence and AI in
  Games}, 4(1):1--49, 2012.

\bibitem{browne09}
C.~B. Browne.
\newblock {\em Automatic Generation and Evaluation of Recombination Games}.
\newblock PhD thesis, Queensland University of Technology, 2009.

\bibitem{Cazenave2020Polygames}
T.~Cazenave, Y.-C. Chen, G.~Chen, S.-Y. Chen, X.-D. Chiu, J.~Dehos, M.~Elsa,
  Q.~Gong, H.~Hu, V.~Khalidov, C.-L. Li, H.-I. Lin, Y.-J. Lin, X.~Martinet,
  V.~Mella, J.~Rapin, B.~Roziere, G.~Synnaeve, F.~Teytaud, O.~Teytaud, S.-C.
  Ye, Y.-J. Ye, S.-J. Yen, and S.~Zagoruyko.
\newblock Polygames: Improved zero learning.
\newblock {\em ICGA Journal}, 2020.
\newblock To appear.

\bibitem{Coulom2007}
R.~Coulom.
\newblock Efficient selectivity and backup operators in {M}onte-{C}arlo tree
  search.
\newblock In H.~J. van~den Herik, P.~Ciancarini, and H.~H. L.~M. Donkers,
  editors, {\em Computers and Games}, volume 4630 of {\em LNCS}, pages 72--83.
  Springer Berlin Heidelberg, 2007.

\bibitem{Cox2009Factoring}
E.~Cox, E.~Schkufza, R.~Madsen, and M.~R. Genesereth.
\newblock In {\em Proceedings of the IJCAI Workshop on General Intelligence in
  Game-Playing Agents (GIGA)}, pages 13--20, 2009.

\bibitem{Emslie2019GalvaniseZero}
R.~Emslie.
\newblock Galvanise zero.
\newblock https://github.com/richemslie/galvanise\_zero, 2019.

\bibitem{Goldwaser2020DeepRLGGP}
A.~Goldwaser and M.~Thielscher.
\newblock Deep reinforcement learning for general game playing.
\newblock In {\em The Thirty-Fourth AAAI Conference on Artificial
  Intelligence}, pages 1701--1708. AAAI Press, 2020.

\bibitem{Kocsis2006BanditBased}
L.~Kocsis and C.~Szepesv{\'a}ri.
\newblock Bandit based {M}onte-{C}arlo planning.
\newblock In J.~F{\"u}rnkranz, T.~Scheffer, and M.~Spiliopoulou, editors, {\em
  Machine Learning: ECML 2006}, volume 4212 of {\em LNCS}, pages 282--293.
  Springer, Berlin, Heidelberg, 2006.

\bibitem{kowalski19}
J.~Kowalski, M.~Maksymilian, J.~Sutowicz, and M.~Szyku{\l}a.
\newblock Regular boardgames.
\newblock In {\em The Thirty-Third AAAI Conference on Artificial Intelligence},
  pages 1699--1706. AAAI Press, 2019.

\bibitem{LanctotEtAl2019OpenSpiel}
M.~Lanctot, E.~Lockhart, J.-B. Lespiau, V.~Zambaldi, S.~Upadhyay,
  J.~P\'{e}rolat, S.~Srinivasan, F.~Timbers, K.~Tuyls, S.~Omidshafiei,
  D.~Hennes, D.~Morrill, P.~Muller, T.~Ewalds, R.~Faulkner, J.~Kram\'{a}r,
  B.~de~Vylder, B.~Saeta, J.~Bradbury, D.~Ding, S.~Borgeaud, M.~Lai,
  J.~Schrittwieser, T.~Anthony, E.~Hughes, I.~Danihelka, and J.~Ryan-Davis.
\newblock {OpenSpiel}: A framework for reinforcement learning in games.
\newblock http://arxiv.org/abs/1908.09453, 2019.

\bibitem{LeCun2015}
Y.~LeCun, Y.~Bengio, and G.~Hinton.
\newblock Deep learning.
\newblock {\em Nature}, 521(7553):436--444, 2015.

\bibitem{LeCun1989CNNs}
Y.~LeCun, B.~Boser, J.~S. Denker, D.~Henderson, R.~E. Howard, W.~Hubbard, and
  L.~D. Jackel.
\newblock Backpropagation applied to handwritten zip code recognition.
\newblock {\em Neural Computation}, 1(4):541--551, 1989.

\bibitem{lin}
M.~Lin, Q.~Chen, and S.~Yan.
\newblock Network in network, 2014.

\bibitem{love08}
N.~Love, T.~Hinrichs, D.~Haley, E.~Schkufza, and M.~Genesereth.
\newblock General game playing: Game description language specification, 2008.

\bibitem{Paszke2019PyTorch}
A.~Paszke, S.~Gross, F.~Massa, A.~Lerer, J.~Bradbury, G.~Chanan, T.~Killeen,
  Z.~Lin, N.~Gimelshein, L.~Antiga, A.~Desmaison, A.~Kopf, E.~Yang, Z.~DeVito,
  M.~Raison, A.~Tejani, S.~Chilamkurthy, B.~Steiner, L.~Fang, J.~Bai, and
  S.~Chintala.
\newblock Pytorch: An imperative style, high-performance deep learning library.
\newblock In H.~Wallach, H.~Larochelle, A.~Beygelzimer, F.~d\textquotesingle
  Alch\'{e}-Buc, E.~Fox, and R.~Garnett, editors, {\em Advances in Neural
  Information Processing Systems 32}, pages 8024--8035. Curran Associates,
  Inc., 2019.

\bibitem{Petosa2019MultiplayerAlphaZero}
N.~Petosa and T.~Balch.
\newblock Multiplayer alphazero.
\newblock In {\em Workshop on Deep Reinforcement Learning at the 33rd
  Conference on Neural Information Processing Systems (NeurIPS 2019)}, 2019.

\bibitem{Piette2020LudiiGameLogicGuide}
{\'E}.~Piette, C.~Browne, and D.~J. N.~J. Soemers.
\newblock Ludii game logic guide.
\newblock {\em CoRR}, abs/2101.02120, 2021.

\bibitem{Piette2020Ludii}
{\'E}.~Piette, D.~J. N.~J. Soemers, M.~Stephenson, C.~F. Sironi, M.~H.~M.
  Winands, and C.~Browne.
\newblock Ludii -- the ludemic general game system.
\newblock In G.~D. Giacomo, A.~Catala, B.~Dilkina, M.~Milano, S.~Barro,
  A.~Bugarín, and J.~Lang, editors, {\em Proceedings of the 24th European
  Conference on Artificial Intelligence (ECAI 2020)}, volume 325 of {\em
  Frontiers in Artificial Intelligence and Applications}, pages 411--418. IOS
  Press, 2020.

\bibitem{Pitrat68GGP}
J.~Pitrat.
\newblock Realization of a general game-playing program.
\newblock In A.~J.~H. Morrel, editor, {\em Information Processing, Proceedings
  of {IFIP} Congress 1968, Edinburgh, UK, 5-10 August 1968, Volume 2 -
  Hardware, Applications}, pages 1570--1574, 1968.

\bibitem{unets}
O.~Ronneberger, P.~Fischer, and T.~Brox.
\newblock U-net: Convolutional networks for biomedical image segmentation.
\newblock In N.~Navab, J.~Hornegger, W.~M. Wells, and A.~F. Frangi, editors,
  {\em Medical Image Computing and Computer-Assisted Intervention -- MICCAI
  2015}, pages 234--241, Cham, 2015.

\bibitem{Schkufza2008Propositional}
E.~Schkufza, N.~Love, and M.~Genesereth.
\newblock Propositional automata and cell automata: Representational frameworks
  for discrete dynamic systems.
\newblock In W.~Wobcke and M.~Zhang, editors, {\em AI 2008: Advances in
  Artificial Intelligence}, volume 5360 of {\em LNCS}, pages 56--66. Springer,
  Berlin, Heidelberg, 2008.

\bibitem{Schrittwieser2019MuZero}
J.~Schrittwieser, I.~Antonoglou, T.~Hubert, K.~Simonyan, L.~Sifre, S.~Schmitt,
  A.~Guez, E.~Lockhart, D.~Hassabis, T.~Graepel, T.~Lillicrap, and D.~Silver.
\newblock Mastering atari, go, chess and shogi by planning with a learned
  model.
\newblock {\em Nature}, 588:604--609, 2020.

\bibitem{fc}
E.~{Shelhamer}, J.~{Long}, and T.~{Darrell}.
\newblock Fully convolutional networks for semantic segmentation.
\newblock {\em IEEE Transactions on Pattern Analysis and Machine Intelligence},
  39(4):640--651, 2017.

\bibitem{Silver2018AlphaZero}
D.~Silver, T.~Hubert, J.~Schrittwieser, I.~Antonoglou, M.~Lai, A.~Guez,
  M.~Lanctot, L.~Sifre, D.~Kumaran, T.~Graepel, T.~Lillicrap, K.~Simonyan, and
  D.~Hassabis.
\newblock A general reinforcement learning algorithm that masters chess, shogi,
  and {G}o through self-play.
\newblock {\em Science}, 362(6419):1140--1144, 2018.

\bibitem{Silver2017AlphaGoZero}
D.~Silver, J.~Schrittwieser, K.~Simonyan, I.~Antonoglou, A.~Huang, A.~Guez,
  T.~Hubert, L.~Baker, M.~Lai, A.~Bolton, Y.~Chen, T.~Lillicrap, F.~Hui,
  L.~Sifre, G.~van~den Driessche, T.~Graepel, and D.~Hassabis.
\newblock Mastering the game of {G}o without human knowledge.
\newblock {\em Nature}, 550:354--359, 2017.

\bibitem{sironi17}
C.~F. Sironi and M.~H.~M. Winands.
\newblock Optimizing propositional networks.
\newblock In {\em Computer Games}, pages 133--151. Springer, 2017.

\bibitem{greatfast}
D.~J. Wu.
\newblock Accelerating self-play learning in go.
\newblock {\em CoRR}, abs/1902.10565, 2019.

\end{thebibliography}

% or include bibliography directly:
%\begin{thebibliography}{0}
%\bibitem[(Author, Year)]{r1} F. Author, Information about cited object.
%
%\bibitem[(Author and Author, Year)]{r2} S. Author and T. Author, Information about cited object.
%\end{thebibliography}

\end{document}